\documentclass[11pt,letterpaper]{article}
\usepackage{emnlp2017}
\usepackage{times}
\usepackage{latexsym}
\usepackage{graphicx,amsfonts}
\graphicspath{ {/} }
\usepackage[font=small,skip=7pt]{caption}

\emnlpfinalcopy

\def\emnlppaperid{***}

\title{Deep Residual Learning for Weakly-Supervised Relation Extraction}

\author{Yi Yao Huang \\
  Department of Electrical Engineering \\ National Taiwan University \\
  Taipei, Taiwan \\
  {\tt b02901042@ntu.edu.tw} \\\And
  William Yang Wang \\
  Department of Computer Science \\
  University of California, Santa Barbara\\
  Santa Barbara, CA 93106 USA\\
  {\tt william@cs.ucsb.edu} \\}

\date{}

\begin{document}

\maketitle

\begin{abstract}
Deep residual learning (ResNet)~\cite{he2016deep} is a new method for training very deep neural networks using identity mapping for shortcut connections. ResNet has won the ImageNet ILSVRC 2015 classification task, and achieved state-of-the-art performances in many computer vision tasks. However, the effect of residual learning on noisy natural language processing tasks is still not well understood. In this paper, we design a novel convolutional neural network (CNN) with residual learning, and investigate its impacts on the task of distantly supervised noisy relation extraction. In contradictory to popular beliefs that ResNet only works well for very deep networks, we found that even with 9 layers of CNNs, using identity mapping could significantly improve the performance for distantly-supervised relation extraction.
\end{abstract}

\section{Introduction}

Relation extraction is the task of predicting attributes and relations for entities in a sentence~\cite{zelenko2003kernel,bunescu2005subsequence,guodong2005exploring, DBLP:journals/corr/Yu0HSXZ17}. For example, given a sentence \emph{``\textbf{Barack Obama} was born in \textbf{Honolulu}, Hawaii.''}, a relation classifier aims at predicting the relation of \emph{``\textbf{bornInCity}''}.
Relation extraction is the key component for building relation knowledge graphs, and it is of crucial significance to natural language processing applications such as structured search, sentiment analysis, question answering, and summarization.

A major issue for relation extraction is the lack of labeled training data. In recent years, distant supervision~\cite{mintz2009distant,hoffmann2011knowledge,surdeanu2012multi} emerges as the most popular method for relation extraction--- it uses knowledge base facts to select a set of noisy instances from unlabeled data. Among all the machine learning approaches for distant supervision,
 the recently proposed Convolutional Neural Networks (CNNs) model~\cite{zeng2014relation} achieved the state-of-the-art performance. Following their success,
 Zeng et al.~\shortcite{zeng2015distant} proposed a piece-wise max-pooling strategy to improve the CNNs.
 Various attention strategies~\cite{lin2016attention,shen-huang:2016:COLING} for CNNs are also proposed, obtaining impressive results. However, most of these neural relation extraction models are relatively shallow CNNs---typically only one convolutional layer and one fully connected layer were involved, and it was not clear whether deeper models could have benefits on distilling signals from noisy inputs in this task.

 In this paper, we investigate the effects of training deeper CNNs for distantly-supervised relation extraction. More specifically, we designed a convolutional neural network based on residual learning~\cite{he2016deep}---we show how one can incorporate word embeddings and position embeddings into a deep residual network, while feeding identity feedback to convolutional layers for this noisy relation prediction task. Empirically, we evaluate on the NYT-Freebase dataset~\cite{riedel2010modeling}, and demonstrate the state-of-the-art performance using deep CNNs with identify mapping and shortcuts. In contrast to popular beliefs in vision that deep residual network only works for very deep CNNs, we show that even with a moderately deep CNNs, there are substantial improvements over vanilla CNNs for relation extraction. Our contributions are three-fold:
 \begin{itemize}
     \item We are the first to consider deeper convolutional neural networks for weakly-supervised relation extraction using residual learning;
     \item We show that our deep residual network model outperforms CNNs by a large margin empirically, obtaining state-of-the-art performances;
     \item Our identity mapping with shortcut feedback approach can be easily applicable to any variants of CNNs for relation extraction.
 \end{itemize}

\begin{figure*}[t!]
\centering
\includegraphics[width=0.8\textwidth]{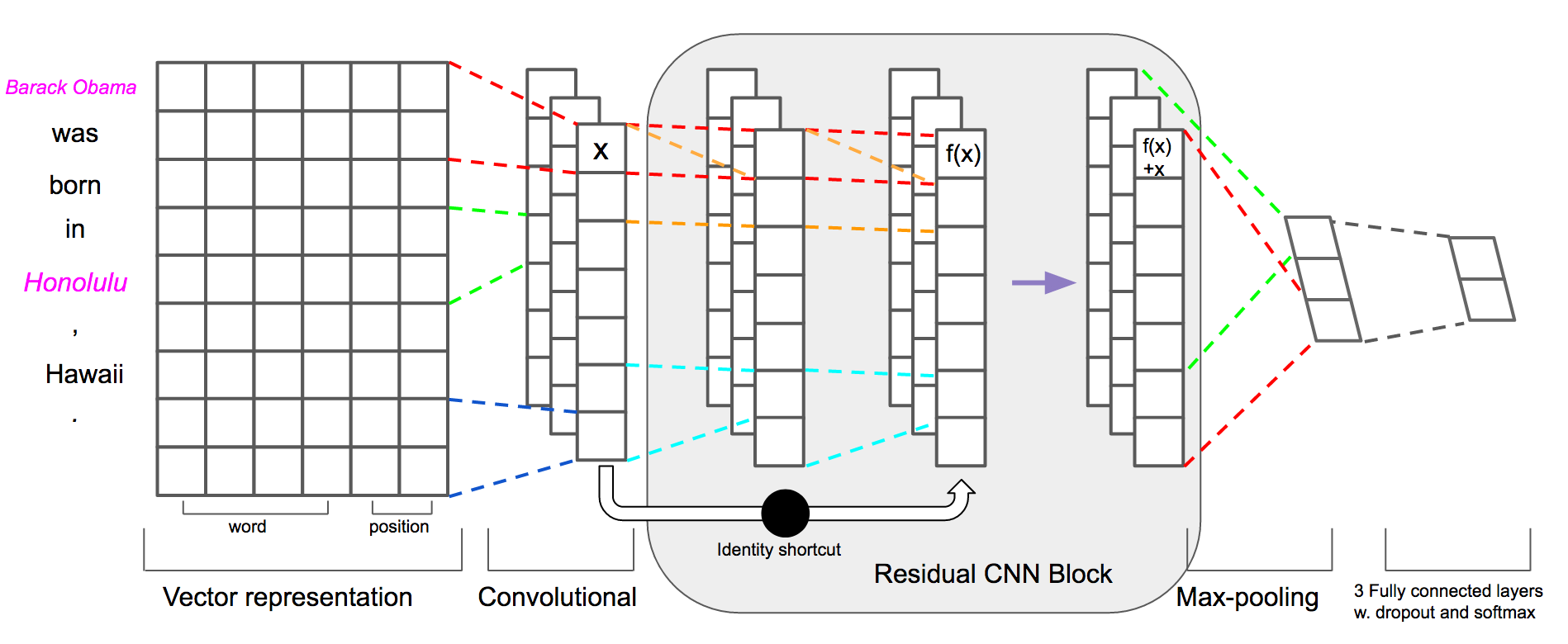}
\caption{The architecture of ResCNN used for relation extraction.}
\label{fig:arch}
\vspace{-2ex}
\end{figure*}

\section{Deep Residual Networks for Relation Extraction}

In this section, we describe a novel deep residual learning architecture for distantly supervised relation extraction. Figure~\ref{fig:arch} describes the architecture of our model.
\subsection{Vector Representation}
Let $\textbf{x}_i$ be the \textit{i}-th word in the sentence and \textit{e1}, \textit{e2} be the two corresponding entities. Each word will access two embedding look-up tables to get the word embedding $\textbf{WF}_i$ and the position embedding $\textbf{PF}_i$. Then, we concatenate the two embeddings and denote each word as a vector of $\textbf{v}_i = [\textbf{WF}_i, \textbf{PF}_i]$.
\subsubsection{Word Embeddings}
Each representation $\textbf{v}_i$ corresponding to $\textbf{x}_i$ is a real-valued vector. All of the vectors are encoded in an embeddings matrix $\textbf{V}_w \in \mathbb{R}^{{d_w} \times |V|}$ where $V$ is a fixed-sized vocabulary.
\subsubsection{Position Embeddings}
In relation classification, we focus on finding a relation for entity pairs. Following \cite{zeng2014relation}, a PF is the combination of the relative distances of the current word to the first entity $\textit{e}_1$ and the second entity $\textit{e}_2$. For instance, in the sentence "\textit{Steve\_Jobs is the founder of Apple.}", the relative distances from \textit{founder} to $\textit{e}_1$ (\textit{Steve\_Job}) and $\textit{e}_2$ are 3 and -2, respectively. We then transform the relative distances into real valued vectors by looking up one randomly initialized position embedding matrices \(\textbf{V}_p \in \mathbb{R}^{{d_p}\times \|P\|}\) where P is fixed-sized distance set. It should be noted that if a word is too far from entities, it may be not related to the relation. Therefore, we choose maximum value $\textit{e}_{max}$ and minimum value $\textit{e}_{min}$ for the relative distance.

In the example shown in Figure~\ref{fig:arch}, it is assumed that $\textit{d}_w$ is 4 and $\textit{d}_p$ is 1. There are two position embeddings: one for $\textit{e}_1$, the other for $\textit{e}_2$. Finally, we concatenate the word embeddings and position embeddings of all words and denote a sentence of length \textit{n} (padded where necessary) as a vector
\vspace{-1ex}
\[
\textbf{v} = \textbf{v}_1 \oplus \textbf{v}_2 \oplus ... \oplus \textbf{v}_n
\vspace{-1ex}
\]
where $\oplus$ is the concatenation operator and \(\textbf{v}_i \in \mathbb{R}^{d}\) (\(d = d_w+d_p\times 2\)).

\subsection{Convolution}
Let $\textbf{v}_{i:i+j}$ refer to the concatenation of words $\textbf{v}_i,\textbf{v}_{i+1},...,\textbf{v}_{i+j}$. A convolution operation involves a \textit{filter} \(\textbf{w}\in\mathbb{R}^{hd}\), which is applied to a window of \textit{h} words to produce a new feature. A feature $\textit{c}_i$ is generated from a window of word $\textbf{v}_{i:i+h-1}$ by
\vspace{-1ex}
\[
\textit{c}_i = \textit{f}(\textbf{w}\cdot\textbf{x}_{i:i+h-1}+\textit{b})
\vspace{-1ex}
\]
Here $\textit{b} \in \mathbb{R}$ is a bias term and \textit{f} is a non-linear function. This filter is applied to each possible window of words from $\textbf{v}_1$ to $\textbf{v}_n$ to produce \textit{feature}
$\textbf{c}=[c_1,c_2,...,c_{n-h+1}]$
with $\textbf{c} \in \mathbb{R}^s(s=n-h+1)$.
\subsection{Residual Convolution Block}
Residual learning connects low-level to high-level representations directly, and tackles the vanishing gradient problem in deep networks. In our model, we design the residual convolution block by applying shortcut connections. Each residual convolutional block is a sequence of two convolutional layers, each one followed by an ReLU activation. The kernel size of all convolutions is $h$, with padding such that the new feature will have the same size as the original one.  Here are two convolutional \textit{filter} $\textbf{w}_1$, $\textbf{w}_2 \in \mathbb{R}^{h\times1}$. For the first convolutional layer:
\vspace{-1ex}
\[
\textit{$\tilde{c}$}_i = \textit{f}(\textbf{w}_1\cdot\textit{c}_{i:i+h-1}+\textit{b}_1)
\vspace{-1ex}
\]
For the second convolutional layer:
\vspace{-1ex}
\[
\textit{$\acute{c}$}_i = \textit{f}(\textbf{w}_2\cdot\textit{$\tilde{c}$}_{i:i+h-1}+\textit{b}_2)
\vspace{-1ex}
\]
Here $\textit{b}_1$, $\textit{b}_2$ are bias terms. For the residual learning operation:
\vspace{-1ex}
\[
\textbf{c} = \textbf{c} + \acute{\textbf{c}}
\vspace{-1ex}
\]
Conveniently, the notation of \textbf{c} on the left is changed to define as the output vectors of the block. This operation is performed by a shortcut connection and element-wise addition. This block will be multiply concatenated in our architecture.
\subsection{Max Pooling, Softmax Output}
We then apply a max-pooling operation over the \textit{feature} and take the maximum value \(\hat{c}=max\{\textbf{c}\}\). We have described the process by which \textit{one} feature is extracted from \textit{one} filter. Take all features into one high level extracted feature \(\textbf{z}=[\hat{c}_1,\hat{c}_2,...,\hat{c}_m]\)(note that here we have \textit{m} filters). Then, these features are passed to a fully connected softmax layer whose output is the probability distribution over relations. Instead of using $y=\textbf{w}\cdot\textbf{z}+b$
for output unit \textit{y} in forward propagation, dropout uses
$y=\textbf{w}\cdot(\textbf{z}\circ\textbf{r})+b$
where $\circ$ is the element-wise multiplication operation and $\textbf{r}\in\mathbb{R}^m$ is a 'masking' vector of Bernoulli random variables with probability \textit{p} of being 1. In the test procedure, the learned weight vectors are scaled by \textit{p} such that \(\hat{\textbf{w}}=\textit{p}\textbf{w}\) and used (without dropout) to score unseen instances.
\section{Experiments}

\subsection{Experimental Settings}
In this paper, we use the word embeddings released by \cite{lin2016attention} which are trained on the NYT-Freebase corpus~\cite{riedel2010modeling}.
We fine tune our model using validation on the training data. The word embedding is of size 50. The input text is padded to a fixed size of 100.  Training is performed with tensorflow adam optimizer, using a mini-batch of size 64, an initial learning rate of 0.001. We initialize our convolutional layers following \cite{glorot2010understanding}. The implementation is done using Tensorflow 0.11. All experiments are performed on a single NVidia Titan X (Pascal) GPU. In Table \ref{table:parameter} we show all parameters used in the experiments.

\begin{table}[h!]
\small
\centering
\begin{tabular}{ |c|c| }
 \hline
 Window size \textit{h} & 3 \\
 Word dimension $\textit{d}_w$ & 50 \\
 Position dimension $\textit{d}_p$ & 5 \\
 Position maximum distance $\textit{e}_{max}$ & 30 \\
 Position minimum distance $\textit{e}_{min}$ & -30 \\
 Number of filters \textit{m} & 128 \\
 Batch size \textit{B} & 64 \\
 Learning rate $\lambda$ & 0.001 \\
 Dropout probability \textit{p} & 0.5 \\
 \hline
\end{tabular}
\caption{Parameter settings}
\vspace{-2ex}
\label{table:parameter}
\end{table}

\noindent We experiment with several state-of-the-art baselines and variants of our model.
\begin{itemize}
  \item \textbf{CNN-B}: Our implementation of the CNN baseline~\cite{zeng2014relation} which contains one convolutional layer, and one fully connected layer. 
  \item \textbf{CNN+ATT}: CNN-B with attention over instance learning \cite{lin2016attention}.
  \item \textbf{PCNN+ATT}: Piecewise CNN-B with attention over instance learning  \cite{lin2016attention}.
  \item \textbf{CNN}: Our CNN model which includes one convolutional layer and three fully connected layers.
  \item \textbf{CNN-x}: Deeper CNN model which has x convolutional layers. For example, CNN-9 is a model constructed with 9 convolutional layers (1 + 4 residual cnn block without identity shortcut) and three fully connected layers.
  \item \textbf{ResCNN-x}: Our proposed CNN-x model with residual identity shortcuts.
\end{itemize}
We evaluate our models on the widely used NYT freebase larger dataset~\cite{riedel2010modeling}. Note that ImageNet dataset used by the original ResNet paper~\cite{he2016deep} has 1.28 million training instances. NYT freebase dataset includes 522K training sentences, which is the largest dataset in relation extraction, and it is the only suitable dataset to train deeper CNNs.
\subsection{NYT-Freebase Dataset Performance}
The advantage of this dataset is that there are 522,611 sentences in training data and 172,448 sentences in testing data and this size can support us to train a deep network. 
Similar to previous work \cite{zeng2015distant,lin2016attention}, we evaluate our model using the held-out evaluation. We report both the aggregate curves precision/recall curves and Precision@N (P@N).
\begin{figure}[t]
\centering
\includegraphics[width=0.5\textwidth]{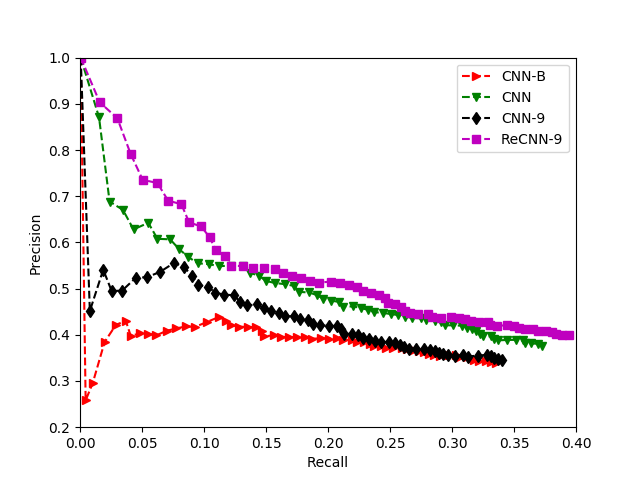}
\caption{Comparing ResCNN to different CNNs.}
\label{fig:CNN_ResCNN}
\vspace{-2ex}
\end{figure}

In Figure \ref{fig:CNN_ResCNN}, we compare the proposed ResCNN model with various CNNs. First, CNNs with multiple fully-connected layers obtained very good results, which is a novel finding. Second, the results also suggest that deeper CNNs with residual learning help extracting signals from noisy distant supervision data. We observe that overfitting happened when we try to add more layers and the performance of CNN-9 is much worse than CNN. We find that ResNet can solve this problem and ResCNN-9 obtains better performance as compared to CNN-B and CNN and dominates the precision/recall curve overall.

\begin{figure}[t]
\begin{center}
\includegraphics[width=0.5\textwidth]{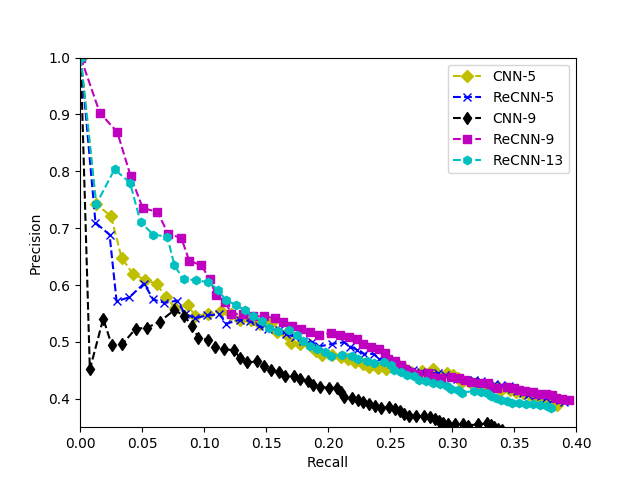}
\caption{Varying the depths of our models.}
\label{fig:DeepCNN}
\vspace{-2ex}
\end{center}
\end{figure}

We show the effect of depth in residual networks in Figure \ref{fig:DeepCNN}. We observe that ResCNN-5 is worse than CNN-5 because the ResNet does not work well for shallow CNNs, and this is consistent with the original ResNet paper. As we increase the network depth, we see that CNN-9 does overfit to the training data.
With residual learning, both ResCNN-9 and ResCNN-13 provide significant improvements over CNN-5 and ResCNN-5 models. In contradictory  to  popular  beliefs  that  ResNet only works well for very deep networks, we found that even with 9 layers of CNNs, using identity mapping could significantly improve the performance learning in a noisy input setting.

\begin{table}[t]
\small
\centering
\begin{tabular}{ |p{1.8cm}||p{0.8cm}|p{0.8cm}|p{0.8cm}|p{0.8cm}| }
 \hline
 P@N(\%)& 100 & 200 & 300 & Mean \\
 \hline
 CNN+ATT  & 76.2  & 68.6 & 59.8 & 68.2  \\
 \hline
 PCNN+ATT  & \textbf{76.2}  & \textbf{73.1} & \textbf{67.4} & \textbf{72.2}  \\
 \hline\hline
 CNN-B  & 41.0  & 40.0 & 41.0 & 40.7  \\
 \hline
 CNN  & 64.0  & 61.0 & 55.3 & 60.1  \\
 \hline
 CNN-5  & 64.0  & 58.5 & 54.3 & 58.9  \\
 \hline
 ResCNN-5  & 57.0  & 57.0 & 54.3 & 56.1  \\
 \hline
 CNN-9  & 56.0  & 54.0 & 49.7 & 53.2  \\
 \hline
 ResCNN-9  & \textbf{79.0}  & \textbf{69.0} & \textbf{61.0} & \textbf{69.7}  \\
 \hline
 ResCNN-13  & 76.0  & 65.0 & 60.3 & 67.1  \\
\hline
\end{tabular}
\caption{P@N for relation extraction with different models. Top: models that select training data. Bottom: models without selective attention.} 
\label{table:1}
\vspace{-2ex}
\end{table}
The intuition of ResNet help this task in two aspect. First, if the lower, middle, and higher levels learn hidden lexical, syntactic, and semantic representations respectively, sometimes it helps to bypass the syntax to connect lexical and semantic space directly. Second, ResNet tackles the vanishing gradient problem which will decrease the effect of noise in distant supervision data.

In Table \ref{table:1}, we compare the performance of our models to state-of-the-art baselines. We show that our ResCNN-9 outperforms all models that do not select training instances. And even without piecewise max-pooling and instance-based attention, our model is on par with the PCNN+ATT model.

For the more practical evaluation, we compare the results for precision@N where N is small (1, 5, 10, 20, 50) in Table \ref{table:2}. We observe that our ResCNN-9 model dominates the performance when we predict the relation in the range of higher probability. ResNet helps CNNs to focus on the highly possible candidate and mitigate the noise effect of distant supervision. We believe that residual connections actually can be seen as a form of renormalizing the gradients, which prevents the model from overfitting to the noisy distant supervision data.

\begin{table}[t]
\small
\centering
\begin{tabular}{ |p{1.8cm}||p{0.6cm}|p{0.6cm}|p{0.6cm}|p{0.6cm}|p{0.6cm}| }
 \hline
 P@N(\%)& 1 & 5 & 10 & 20 & 50 \\
 \hline
 PCNN+ATT  & \textbf{1}  & 0.8 & \textbf{0.9} & 0.75 & 0.7 \\
 \hline
 ResCNN-9  & \textbf{1}  & \textbf{1} & \textbf{0.9} & \textbf{0.9} & \textbf{0.88} \\
 \hline
\end{tabular}
\caption{P@N for relation extraction with different models where N is small. We get the result of PCNN+ATT using their public source code.}
\label{table:2}
\vspace{-2ex}
\end{table}

In our distant-supervised relation extraction experience, we have two important observations:
(1) We get significant improvements with CNNs adding multiple fully-connected layers. (2) Residual learning could significantly improve the performance for deeper CNNs.


\section{Conclusion}
In this paper, we introduce a deep residual learning method for distantly-supervised relation extraction. We show that
deeper convolutional models help distill signals from noisy inputs. With shortcut connections and identify mapping, the performances are significantly improved. These results
aligned with a recent study~\cite{conneau2016very},
suggesting that deeper CNNs do have positive effects on noisy NLP problems.

\bibliography{all}
\bibliographystyle{emnlp_natbib}

\end{document}